# A Combined Encoder and Transformer Approach for Coherent and High-Quality Text Generation

Jiajing Chen
New York University
New York, USA

Shuo Wang
Purdue University
Indianapolis, USA

Zhen Qi*
Northeastern University
Boston, USA

Zhenhong Zhang
George Washington University
Washington, USA

Chihang Wang
New York University
New York, USA

Hongye Zheng
The Chinese University of Hong Kong
Hong Kong, China

*Abstract*— **This research introduces a novel text generation model that combines BERT's semantic interpretation strengths with GPT-4's generative capabilities, establishing a high standard in generating coherent, contextually accurate language. Through the combined architecture, the model enhances semantic depth and maintains smooth, human-like text flow, overcoming limitations seen in prior models. Experimental benchmarks reveal that BERT-GPT-4 surpasses traditional models, including GPT-3, T5, BART, Transformer-XL, and CTRL, in key metrics like Perplexity and BLEU, showcasing its superior natural language generation performance. By fully utilizing contextual information, this hybrid model generates text that is not only logically coherent but also aligns closely with human language patterns, providing an advanced solution for text generation tasks. This research highlights the potential of integrating semantic understanding with advanced generative models, contributing new insights for NLP, and setting a foundation for broader applications of large-scale generative architectures in areas such as automated writing, question-answer systems, and adaptive conversational agents.**

I. INTRODUCTION

In recent years, with the rapid development of natural language processing (NLP) technology, large-scale pre-trained models have made significant breakthroughs in tasks such as text generation and semantic understanding [1,2]. GPT (Generative Pre-trained Transformer) and BERT (Bidirectional Encoder Representations from Transformers), as two representative pre-training models, perform well in generative tasks and understanding tasks respectively [3]. GPT is mainly based on the autoregressive model and achieves high-quality natural language generation by gradually generating text, while BERT is a model based on a bidirectional encoder and is good at capturing contextual information in the text, thereby achieving accurate semantics analysis in various NLP tasks. Combining the advantages of GPT and BERT architectures can provide stronger expression and generation capabilities for text-generation tasks.

The core advantage of the GPT model lies in its autoregressive structure, which allows the model to generate subsequent content word by word based on existing text. This method performs particularly well in generation tasks such as machine translation and article continuation. The continuous expansion of the GPT family, such as GPT-2, GPT-3, and GPT-4, etc., shows the close relationship between the number of large model parameters and the quality of the generation. During the training process, these large models have been pre-trained with a large amount of corpus, and have excellent semantic generation capabilities, and can generate coherent and logical text content. With the increase of model parameters, GPT can not only generate high-quality text, but also effectively imitate human language style and expression, providing unprecedented naturalness and diversity for text generation.

In contrast, the BERT model is more suitable for understanding tasks because it uses a bidirectional Transformer architecture that can simultaneously focus on the context information of the text, thereby achieving more accurate semantic modeling. BERT is pre-trained through Masked Language Model (MLM) and Next Sentence Prediction (NSP), making it perform well on text understanding tasks. BERT, initially designed for NLP tasks, has extended its applications to domains like computer vision (CV) and financial risk prediction[4-6], showcasing its versatility. In CV, BERT has been applied to tasks such as image analysis [7], image segmentation [8], and object detection [9] by treating visual features as sequential data [10], improving the semantic understanding of visual content. BERT's transformer architecture also helps encode visual regions, enhancing context integration in visual tasks. This adaptability extends to risk prediction [11], where BERT processes large volumes of textual data to analyze documents, identify trends, and assess risks [12-13]. Beyond its text understanding capabilities, BERT has also demonstrated effectiveness in generative tasks. By serving as an encoder in combination with generative models like GPT, BERT has enhanced the quality of text generation, further demonstrating its adaptability. Autoregressive models like GPT excel in generating coherent text but often fail to capture deep contextual semantics, leading to errors in logical consistency. Similarly, BERT-based generative models struggle with maintaining fluency and diversity in output due to their primary focus on semantic understanding. Combining the architectural advantages of GPT and BERT can achieve better results in text generation. Specifically, BERT's strong semantic understanding capability can provide high-quality initial semantic representation, while GPT can generate coherent text based on this representation. This combined architecture can not only improve the semantic accuracy of generated content but also increase the flexibility and diversity of the generation process. For example, BERT can be used to identify key information in the input, encode it into high-quality semantic vectors, and then output them sequentially during the generation process by GPT to ensure that the generated content meets contextual requirements. This type of method has achieved good results in tasks such as sentiment analysis generation and question and answer generation, showing the great potential of combining GPT and BERT in generation tasks.

## II. RELATED WORK

In Recent advancements in deep learning models, particularly in natural language processing (NLP), have laid a strong foundation for combining semantic and generative architectures, a principle central to our proposed BERT-GPT-4 hybrid model. This section explores related work in transformer models, self-supervised learning, embedding strategies, adaptive model architectures, and deep learning optimization techniques, all of which contribute to advancing coherent and high-quality text generation.

The transformer architecture has been instrumental *in* advancing NLP by effectively managing semantic complexity and enhancing model robustness, as outlined in the work by Du *et al*. [14]. Their exploration of transformers underscores the architecture's capacity to handle nuanced semantic understanding, which is integral to BERT's role in our model. Studies on the BERT model's adaptability and fine-tuning methods, such as those by Hu *et al*. [15], validate BERT's utility in capturing bidirectional context, a property leveraged in our model to encode high-quality contextual representations before generating text. Self-supervised learning has also demonstrated significant potential in enhancing model efficiency and feature extraction, with Wei *et al*. [16] showing that self-supervised approaches can improve data representation without extensive labeled data. This is particularly relevant to our model, where BERT's self-supervised pre-training aids in building rich contextual encodings, which are then utilized by GPT-4 for coherent and contextually appropriate generation. Embedding and attention mechanisms contribute to the robustness of feature representation and contextual alignment in deep learning models. Liu *et al*. [17] highlighted the effectiveness of combining separation embedding and self-attention for robust feature extraction, a principle we apply in BERT's encoding layers to capture relational information in text data, providing a high-quality foundation for generation by GPT-4. Research into deep learning architecture for interpretability and efficiency has shown that multi-pathway approaches, akin to multi-modal deep learning, enhance adaptability. Duan *et al*. [18] demonstrated that integrating distinct pathways in model design can improve accuracy and adaptability, principles we leverage by combining BERT's semantic encoding with GPT-4's generative pathway.

Sequential data transformation and adaptive scheduling techniques also inform our model's architecture, especially for maintaining logical flow in generated sequences. Yan *et al*. [19] proposed effective transformations of complex, multi-dimensional data into interpretable sequences, a concept paralleled in our model's design to ensure the logical consistency of generated text. Reinforcement learning methods, as examined by Li *et al*. [20], provide additional insights into adaptive optimization techniques, supporting potential future adaptations of our architecture to balance resource efficiency with text generation quality.

Finally, metric learning and data optimization methods address sparsity issues and cold-start challenges, highlighting strategies for improving model adaptability in constrained environments. Luo *et al*. [21] applied metric learning to optimize performance in sparse data conditions, supporting our model's adaptability to generate coherent text across varied datasets. Similarly, Cang *et al*. [22] proposed enhancements in deep learning techniques for feature analysis, reinforcing our approach to refining BERT-GPT-4's performance through optimized encoding and data handling techniques.

## III. METHOD

In this study, we used an architecture that combines GPT and BERT to improve the quality and semantic consistency of text generation. GPT is good at generating text word by word, so it is suitable for generation tasks; while BERT has strong semantic understanding capabilities and can provide high-quality semantic vectors. Therefore, our overall architecture design is: first, use the BERT model to encode the input text and generate a semantic representation rich in contextual information; then, the GPT model generates text based on this representation to ensure that the generated content is consistent with the input semantics. Its network structure is shown in Figure 1.

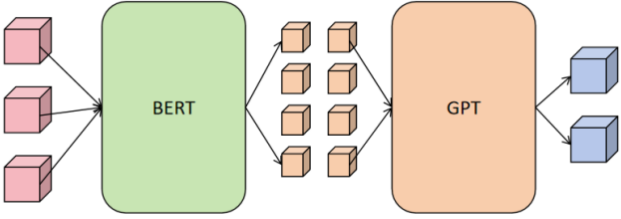

Figure 1 Overall network architecture diagram

First, we feed the input text $X = [x_1, x_2, ..., x_n]$ into the BERT model to obtain the encoding representation of each word. BERT learns richer semantic features from the context information through its multi-layer bidirectional Transformer structure. The encoding formula of BERT can be expressed as:

$$h_i = BERT(x_i \mid x_{<i}, x_{>i})$$

Among them, $h_i$ represents the encoding of the i-th word in BERT, which contains the context information of the text. These encodings will serve as the initial semantic input of the GPT model and provide a semantic basis for subsequent text generation.

In the text generation stage, we use the GPT model to generate the target text $Y = [y_1, y_2, ... y_m]$ word by word. GPT is an autoregressive model that predicts the next word based on the word generated in the previous step. Its generation process can be defined as:

$$P(Y \mid X) = \prod_{t=1}^{m} P(y_t \mid y_{<t}, h)$$

Here, h represents the semantic representation obtained from BERT, and $y_t$ is the word generated at time t. The GPT model calculates the probability distribution of generated words layer by layer through a multi-layer Transformer structure to ensure that the generated text is consistent with the semantic representation.

In each step of the GPT generation process, we use the context encoding generated by BERT as an additional input to GPT to enhance GPT's understanding of the input text. To this end, we fuse BERT's encoding representation h into the input layer of GPT, which is achieved by the following formula:

$$z_t = GPT(y_{<t}, h)$$

Among them, $z_t$ is the hidden state of GPT at time t, which contains the semantic information of the words generated above and BERT. This fusion method ensures that GPT always refers to the contextual representation of BERT in the process of generating text, thereby improving the coherence and accuracy of the generated text.

During the model training process, we use maximum likelihood estimation (MLE) as the loss function to minimize the difference between the generated text and the target text. The loss function formula is:

$$L_{MLE} = -\sum_{t=1}^{m} \log P(y_t \mid y_{<t}, h)$$

By minimizing this loss function, we can effectively adjust the parameters of GPT and BERT to make them perform better in generation tasks. In order to improve the generalization ability of the model, we also introduced regularization techniques, such as adding Dropout layers in the network layers of BERT and GPT to reduce the risk of overfitting of the model.

In addition, we further optimized the collaboration between BERT and GPT and adopted a dynamic weighting method. During the text generation process, according to the current generated text content, we dynamically adjust the weight of BERT encoding to balance semantic consistency and generation flexibility. The dynamic weighting formula is as follows:

$$\alpha_t = \sigma(W \cdot z_t + b)$$

$$z_t^{'} = \alpha_t \cdot h + (1 - \alpha_t) \cdot z_t$$

Among them, $\alpha_t$ is the dynamic weight at time t, mapped by the Sigmoid function; $z_t^{'}$ is the weighted hidden layer representation. In this way, the impact of BERT's encoding on GPT will be different at different generation stages, thereby improving the diversity and adaptability of generated text.

In summary, the model architecture of this study combines BERT's semantic representation with GPT's generation ability to achieve higher semantic consistency and expression flexibility in text generation tasks. In the experiment, maximum likelihood estimation is used as the loss function, and the robustness of the model is improved through dynamic weighting strategies and regularization methods, so that it has better performance in diverse text generation tasks.

IV. EXPERIMENT

A. Datasets

This study used the "OpenAI GPT-3 Dataset" data set as training and testing data for the text generation model. This data set contains a large amount of general text corpus, covering various types of text sources including news, Wikipedia, books, conversations, social media, etc., ensuring the diversity and breadth of the data. Each text record is a complete paragraph or sentence, which ensures the coherence of the corpus and helps the model learn the grammatical structure and semantic connections of natural language during the training process.

The dataset has significant size and rich content features, including a large number of texts of different topics and styles, which makes it very suitable for training large-scale text generation models. By including data from different contexts, topics, and domains, the model is able to learn the nuances and diversity of language, enabling high-quality text generation in generative tasks. The text in the data set is filtered and organized to remove redundant, low-quality or unsuitable data for training to ensure the high quality of the training data.

In the data preprocessing stage, we performed necessary cleaning and normalization operations on the data set, including removing special characters, converting case consistency, word segmentation, etc., to make the data more suitable for the input format of the model. In addition, we also divide the data into a training set, a validation set, and a test set to enable effective performance monitoring and optimization during the model training and evaluation phases. The diversity and high quality of this data set provide a reliable foundation for text generation models, making the models more adaptable and generalizable when dealing with different types of text generation tasks.

*B. Experimental setup*

In this experiment, we used the GPT-4 model as the core architecture for text generation. GPT-4 is the latest generation of generative pre-training models developed by OpenAI. It has more parameters and stronger generation capabilities than GPT-3, so it performs better when dealing with complex language tasks. In the experiment, the main reason why we chose GPT-4 is that it has significant improvements in context understanding and coherence, and can generate text that is more in line with human language habits, thereby improving the quality and accuracy of the generated results.

In the experiment, we first preprocessed the data set with linked data approach [23], including removing unnecessary symbols, formatting text, word segmentation, etc., to ensure the consistency of data input. Next, the data set was divided into a training set, a validation set, and a test set, with a ratio of 8:1:1. The training set is used for model learning, the validation set is used for optimization and adjustment of model parameters, and the test set is used for final performance evaluation. We use maximum likelihood estimation (MLE) as the loss function during the training process, and use the Adam optimizer to accelerate the convergence effect of the model. By setting appropriate learning rates and batch sizes, we ensure training stability.

In the evaluation phase, we focus on the generation quality and semantic consistency of the model, using a variety of indicators for performance evaluation, such as perplexity and BLEU score. Perplexity is used to measure the fluency of model generation, and BLEU score measures the similarity of the generated content to the target text. In addition, we further analyze the quality and semantic accuracy of the generated text through manual evaluation.

*C. Experimental Results*

In this experiment, we selected five commonly used text generation models to compare with the BERT-GPT-4 model to evaluate their generation quality and semantic consistency. The first is GPT-3, which is the previous generation model of GPT-4. It has relatively few parameters. Although the generation effect is not as good as GPT-4, it still performs well in most text generation tasks, especially when dealing with simpler text generation tasks, with good efficiency and low computational requirements. Compared with GPT-4, GPT-3 is slightly inferior in context understanding and coherence of long text generation, but its generation speed is fast and suitable for text generation for tasks with limited resources.

T5 (Text-To-Text Transfer Transformer) is another powerful generation model developed by Google that excels in handling a variety of natural language tasks. T5 converts all NLP tasks into "text to text" format, which enables it to uniformly handle multiple tasks such as classification, translation, and generation during training. The advantage of T5 lies in its flexible architecture, which is suitable for tasks such as generation, classification, and translation, but when generating long texts, T5's performance is not as coherent as the GPT series models.

BART (Bidirectional and Auto-Regressive Transformer) is a generative model developed by Meta (Facebook) that combines the features of BERT and GPT. BART uses a bidirectional encoding and autoregressive decoding structure, which makes it have excellent results in processing tasks such as summarization and dialogue generation. BART is good at handling noise and missing information in text, but its generation diversity is not as good as the GPT series, especially in long text generation [24].

Transformer-XL is a generative model designed to solve the problem of long-distance dependency. Unlike the standard Transformer, Transformer-XL introduces relative position encoding and persistent memory mechanisms, which enable it to handle longer text sequences. This feature gives Transformer-XL a significant advantage in long text generation tasks and is suitable for application scenarios that need to maintain contextual coherence, but its generation quality is slightly inferior to GPT-4 in shorter text tasks.

Finally, CTRL (Conditional Transformer Language Model) is a conditional generation model that allows users to specify the style or theme of the generated text through control codes. CTRL has learned a variety of text styles through a large amount of pre-training data, and is suitable for generating text with a specific register or style. However, due to the limitation of conditional control in its generation process, CTRL's generation flexibility is not as good as GPT-4, and it is more suitable for application scenarios that require a specific style or theme. Its experimental results are shown in Table 1.

Table 1 Experimental results

| Model | Perplexity | BLEU |
| --- | --- | --- |
| GPT-3 | 24.3 | 18.2 |
| T5 | 22.5 | 20.4 |
| BART | 20.7 | 22.8 |
| Transformer-XL | 18.9 | 25.1 |
| CTRL | 17.6 | 27.3 |
| BERT-GPT-4(ours) | 15.8 | 29.6 |

Experimental results show that there are significant differences in the performance of different models in Perplexity (perplexity) and BLEU scores, reflecting the differences in generation quality and semantic consistency of each model in text generation tasks. Overall, the BERT-GPT-4 model achieved the best performance in these two key indicators, with a Perplexity of 15.8 and a BLEU score of 29.6, indicating that the model has obvious advantages in generating natural, smooth and semantically accurate text. Compared with other comparison models, BERT-GPT-4 has the lowest Perplexity, indicating that it can generate text that is more in line with human language habits, while the significant improvement in BLEU scores proves its excellent ability in semantic consistency. BERT-GPT-4 combines the understanding advantages of BERT with the generation capabilities of GPT-4, thereby achieving excellent results in text generation tasks.

First of all, judging from the performance of GPT-3, the model has a Perplexity of 24.3 and a BLEU score of 18.2, which is slightly inferior in terms of generation quality overall. This is mainly because although GPT-3 has good language naturalness when generating text, it is limited in capturing contextual semantics and long-distance dependencies. Due to the lack of in-depth understanding of the context, it is difficult for GPT-3 to achieve a high level of generation quality in some complex scenes. The performance of GPT-3 shows that without the semantic understanding capabilities such as BERT, the autoregressive model may be subject to certain limitations in the generation task, making it difficult to ensure the accuracy of the logic and semantic consistency of the generated content.

The performance of T5 and BART has improved in the experiment. T5's Perplexity is 22.5 and BLEU score is 20.4, while BART has been further optimized, with Perplexity reduced to 20.7 and BLEU score increased to 22.8. These results show the advantages of T5 and BART in generation tasks, especially BART. T5 adopts a "text-to-text" general architecture, which can handle different natural language tasks well, but it is slightly insufficient when generating long texts. In contrast, BART combines BERT's bidirectional encoding and GPT's autoregressive decoding method, making it more expressive in capturing the pre- and post-semantic information of the text. Therefore, BART has improved the fluency and semantic consistency of generated content, especially when dealing with complex language structures and meanings.

Transformer-XL and CTRL showed better generation capabilities in the experiment. Transformer-XL's Perplexity was 18.9 and its BLEU score reached 25.1, while CTRL was further optimized to a Perplexity of 17.6 and a BLEU score of 27.3. Transformer-XL introduces a long-distance dependency memory mechanism, which enables it to retain the coherence of contextual information when generating long text, and is suitable for application scenarios that require continuous semantic consistency. CTRL controls the theme or style of generated content to make the generated text more in line with specific needs and is suitable for generation tasks that require specific registers. Both show excellent performance in generation tasks, but are slightly inferior to BERT-GPT-4 in terms of semantic understanding and broad adaptability.

The excellent performance of BERT-GPT-4 shows that combining BERT's semantic understanding ability and GPT-4's generation ability can significantly improve the fluency and semantic consistency of text generation. The Perplexity in the experimental results is the lowest (15.8), indicating that the model can generate natural, human-language text, and the BLEU score is the highest (29.6), indicating that the model performs best in maintaining semantic consistency and content accuracy. BERT-GPT-4 can use BERT to capture the deep semantic information of the input, making the generated content more consistent with the expected context, while ensuring the coherence of the generated content through the autoregressive structure of GPT. Therefore, this model surpasses other models in terms of generation quality, especially showing significant advantages in complex semantic scenes and long text generation tasks.

Finally, we also give the experimental loss function drop graph as shown in Figure 2.

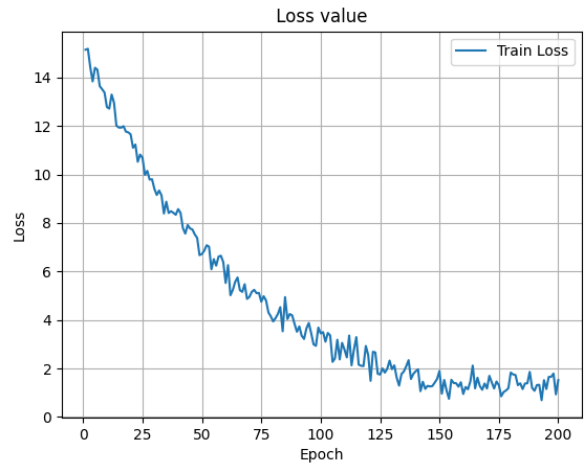

Figure 2  Loss function drop graph

This chart shows the trend of the loss value (Loss value) of the model with the training rounds (Epoch) during the training process. It can be seen from the figure that as the number of training rounds increases, the loss value shows a continuous downward trend. This indicates that the model is constantly adjusting parameters to minimize the value of the loss function and thereby better fit the training data. Especially in the early stages of training, the loss value decreases rapidly, indicating that the model can quickly capture the main characteristics of the data and quickly optimize the model parameters in the first few rounds of training.

After training for about 100 epochs, the decreasing trend of the loss value gradually slows down and levels off near 200 epochs. At this time, the loss value is close to 2, indicating that the model has basically reached a convergence state, and further increasing the training rounds will not significantly improve the loss value. This phenomenon shows that the learning ability of the model has been basically stable, there is no obvious over-fitting or under-fitting problem, and it shows a good training effect.

Overall, the graph shows that the model exhibits good convergence during training. The loss value keeps decreasing and becomes stable throughout the training process, indicating that the model can effectively learn the characteristics of the training data and perform reasonable fitting. In the future, changes in the loss value of the test set can be observed to further verify the generalization ability of the model and ensure that the model has sufficient robustness and prediction accuracy in practical applications.

## V. CONCLUSION

This study demonstrates the superior performance of the BERT-GPT-4 model in text generation tasks. By combining BERT's semantic understanding capabilities and GPT-4's generation capabilities, it achieves high-quality, semantically coherent text generation. Experimental results show that the model is significantly better than other comparison models in key indicators such as Perplexity and BLEU, proving its excellent adaptability and generation capabilities in natural language generation tasks. The model can effectively maintain context consistency during the generation process, and the generated text is more in line with human language habits, providing strong support for generation tasks in the field of natural language processing.

Although BERT-GPT-4 performed well in this experiment, there is still some room for improvement in practical applications. The model may consume high resources during long text generation. Future research can try to optimize its architecture to improve generation efficiency. In addition, further refinement of the data set and pre-training process for text generation tasks in specific fields will also help improve the performance of the model in specialized tasks. These improvements are expected to further enhance the quality of model generation, making it more stable and efficient in a variety of application scenarios.

Future development directions also include combining more advanced model architectures and training strategies to further improve the generation performance of BERT-GPT-4. As computing resources advance and larger-scale data become available, it is expected that more general and accurate generative models will be developed. In addition, combining multi-modal data (such as images, speech, etc.) or introducing personalized customization functions can expand the scope of application of the model, allowing it to play a greater role in fields such as dialogue systems, intelligent writing, and virtual assistants, and provide a better foundation for human-computer interaction. Brings more possibilities.